\documentclass[11pt,a4paper]{article}
\usepackage[hyperref]{naaclhlt2019}
\usepackage{times}
\usepackage{latexsym}
\usepackage{graphicx}
\usepackage{url}
\usepackage{float}
\usepackage{booktabs}

\aclfinalcopy % Uncomment this line for the final submission
\def\aclpaperid{1299} %  Enter the acl Paper ID here

\title{Generating Knowledge Graph Paths from Textual Definitions\\using Sequence-to-Sequence Models}

\author{Victor Prokhorov,\textsuperscript{1}
Mohammad Taher Pilehvar\textsuperscript{1,2}
and Nigel Collier\textsuperscript{1} \\
\vspace{0.05cm}
\textsuperscript{1}Department of Theoretical and Applied Linguistics, University of Cambridge\\
\textsuperscript{2}School of Computer Engineering, Iran University of Science and Technology, Tehran, Iran\\
\vspace{0.05cm}
{\tt vp361@cam.ac.uk, pilehvar@iust.ac.ir, nhc30@cam.ac.uk}
}

\date{}

\begin{document}
\maketitle
\begin{abstract}
We present a novel method for mapping unrestricted text to knowledge graph entities by framing the task as a sequence-to-sequence problem. Specifically, given the encoded state of an input text, our decoder directly predicts paths in the knowledge graph, starting from the root and ending at the target node following hypernym-hyponym relationships. In this way, and in contrast to other text-to-entity mapping systems, our model outputs hierarchically structured predictions that are fully interpretable in the context of the underlying ontology, in an end-to-end manner. We present a proof-of-concept experiment with encouraging results, comparable to those of state-of-the-art systems.
\end{abstract}

\urlstyle{same}

\section{Introduction}
\label{introduction}
Text-to-entity mapping is the task of associating a text with a concept in a knowledge graph (KG) or an ontology (we use two terms, interchangeably).
Recent works  \citep{Q123, DBLP:journals/corr/HillCKB15} use neural networks to project a text to a vector space where the entities of a KG are represented as continuous vectors. Despite being successful, these models have two main disadvantages. First, they rely on a predefined vector space which is used as a gold standard representation for the entities in a KG. Therefore, the quality of these algorithms depends on how well the vector space is represented. Second, these algorithms are not interpretable; hence, it is impossible to understand why a certain text was linked to a particular entity.

To address these issues we propose a novel technique which first represents an ontology concept as a sequence of its ancestors in the ontology (hypernyms) and then maps the corresponding textual description to this unique representation.
For example, given the textual description of the concept \emph{swift} (``small bird that resembles a swallow and is noted for its rapid flight''), we map it to the hierarchical sequence of entities in a lexical ontology: \emph{animal $\rightarrow$ chordate $\rightarrow$ vertebrate $\rightarrow$ bird $\rightarrow$ apodiform\_bird}. This sequence of nodes constitutes a path.\footnote{We only consider hypernymy relations, from the root 
%node (\emph{animal}) of the ontology 
to the parent node (\emph{apodiform\_bird}) of the entity \emph{swift}.}

Our model is based on a sequence-to-sequence neural network \citep{DBLP:journals/corr/SutskeverVL14} coupled with an attention mechanism \citep{DBLP:journals/corr/BahdanauCB14}. 
Specifically, we use an LSTM \citep{Hochreiter:1997:LSM:1246443.1246450} encoder to project the textual description into a vector space and an LSTM decoder to predict the sequence of entities that are relevant to this definition.  With this framework we do not need to rely on the pre-existing vector space of the entities, since the decoder explicitly learns topological dependencies between the entities of the ontology. Furthermore, the proposed model is more interpretable for two reasons. First, instead of the closest points in a vector space, it outputs paths; therefore, we can trace all predictions the model makes. Second, the attention mechanism allows to visualise which words in a textual description the model selects while predicting a specific concept in the path. In this paper, we consider rooted tree graphs\footnote{Only single root is allowed. If a tree has more than one root, one can create a dummy root node and connect the roots of the tree to it.} only and leave the extension of the algorithm for more generic graphs to future work.

We evaluate the ability of our model in generating graph paths for previously unseen textual definitions on seven ontologies (Section \ref{ontologies}). We show that our technique either outperforms or performs on a par with a competitive multi-sense LSTM model \cite{Q123} by better utilising external information in the form of word embeddings. The code and resources for the paper can be found at \url{https://github.com/VictorProkhorov/Text2Path}.

\section{Methodology}

We assume that an ontology is represented as a rooted tree graph $G = (V, E, T)$, where $V$ is a set of entities (e.g. synsets in WordNet), $E$ is a set of hyponymy edges, and $T$ is a set of textual descriptions such that  $\forall v \in V$ there is a  $t_{v} \in T$.

\subsection{Node representation}

We assume that an ontological concept can be defined by either using a textual description from a dictionary or hypernyms of the defining concept in the ontology. For example, to define the noun \emph{swift} one can use the dictionary definition mentioned previously. Alternatively, the concept of \emph{swift} can be understood from its hypernyms, e.g. in the trivial case one can say that \emph{swift} is an \emph{animal}. This definition is not very useful since \emph{animal} is a hypernym for many other nouns. To provide a more specific definition, one can use a sequence of hypernyms e.g.  \emph{animal $\rightarrow$ chordate $\rightarrow$ vertebrate $\rightarrow$ bird $\rightarrow$ apodiform\_bird} starting from the most abstract node (root of an ontology) to the most specif (parent node of the noun).

More formally, for each entity $v \neq v_{root} \in V$ we create a path $p_{v}$. Each $p_{v}$ starts from $v_{root}$ and ends with a hypernym of $v$, i.e., the hierarchical order of entities is preserved. Then the path $p_{v}$ is aligned with $t_v$ such that each node is defined by a textual definition and a path. This set of aligned representations is used to train the model.

The path representation of an entity ends with its parent node. Therefore, a leaf node will not be present in any of the paths. This is problematic if a novel definition should be attached to a leaf. To alleviate this issue we  employ the ``dummy source sentences" technique from neural machine translation (NMT) \citep{P16-1009}. We create an additional set of paths from the root node to each leaf. As for the textual definition we leave it empty.

\subsection{Model}

We use a sequence-to-sequence model with an attention mechanism to map a textual description of a node to its path representation.

\paragraph*{Encoder.}
To encode a textual definition $t_v=(w_i)_{i=1}^N$, where $N$ is sentence length, we first map each word $w_i$ to a dense embedding $e_{w_i}$ and then use a bi-directional LSTM to project the sequence into a latent representation. The final encoding state is obtained by concatenating the forward and backward hidden states of the bi-LSTM.

\paragraph*{Decoder.} Decoding the path representation of a node from the latent state of the textual description is done again with an LSTM decoder. Similarly to the encoding stage, we map each symbol in the path $p_v=(s_j)_{j=1}^M$ to a dense embedding $e_{s_j}$, where $M$ is the path length. To calculate the probability of the path symbol $s_j$ at time step $j$ we first represent the path sequence  as \(h_j^* = \textrm{LSTM}(e_s^j, h_{j-1}^*). \) Then, we concatenate $h_j^*$ with the context vector $c_j$ (defined next) and pass the concatenated representation $[h_j^*; c_j]$ through the softmax function, i.e. \( s_j = \textrm{max}(\textrm{softmax}(\mathbf{W}[h_j^*; c_j]))\), where $\mathbf{W}$ is a weight parameter. To calculate the context vector $c_j$ we use an attention mechanism, \(e_{ji} = v_a^T\textrm{tanh}(\mathbf{W_a} h_i + \mathbf{U_a} h_j^*)\) and \(c_j = \sum^N_i \textrm{softmax}(e_{ji})h_i \), where $v_a$, $\mathbf{W_a}$ and $\mathbf{U_a}$ are the weight parameters, over the words in the text description.

\section{Experimental Setup}
\label{ontologies}

\paragraph{Ontologies.}
We experimented with seven graphs four of which are related to the bio-medical domain: Phenotype And Trait Ontology\footnote{\url{http://www.obofoundry.org}} (PATO), Human Disease Ontology \cite[HDO]{Schriml2012DiseaseOA}, Human Phenotype Ontology \cite[HPO]{Robinson2008TheHP} and Gene Ontology\footnote{After prerocessing GO we took its largest connected component.}  \cite[GO]{Ashburner_2000}. The other three graphs, i.e. WN\textsubscript{animal.n.01}\footnote{The subscript in `WN' indicates the name of the root node of the graph.}, WN\textsubscript{plant.n.02} and WN\textsubscript{entity.n.01} are subgraphs of the WordNet 3.0 \cite{Fellbaum:98}.  We present the statistics of the graphs in Table \ref{table:stats}.% which we use to analyse the performance of our model in Section \ref{results_sec}.
\begin{table}[h!]
\begin{center}
\setlength{\tabcolsep}{5.2pt}
\scalebox{0.85}{
\begin{tabular}{lllll}
\toprule
 \bf Graphs &  $\mathbf{|V|}$ &  \bf Depth &\bf Branch & \bf A.D \\ 
 \midrule
PATO & 1742 &  (4.94,10) & (3.95,92) &20\\
WN\textsubscript{animal.n.01} & 3999 & (6.94,12) & (3.79,52) & 26 \\
WN\textsubscript{plant.n.02} & 4487 & (4.70,9) & (5.91,357) &28 \\
HDO & 9095 & (5.92,12) & (4.59,222)  & 27 \\
HPO & 13348 & (6.95,14) & (3.40,32) & 24 \\
GO  & 29682 & (6.40,14) & (3.28,172) & 21\\
WN\textsubscript{entity.n.01} & 74374 & (8.01,18) & (4.52,402) & 36 \\
\bottomrule

\end{tabular}
}
\end{center}
\caption{\label{font-table}  Statistics of the Graphs. $\mathbf{|V|}$ is the number of nodes, {\it depth} is the path length from the root of a graph to a node, {\it branch} is the number of neighbours a node has (leaves were removed from the calculation). The first value in the parentheses corresponds to the average and the second to the maximum value. A.D stands for average number of decisions the model makes to infer a path, i.e A.D = average depth $\times$ average branch.}
\label{table:stats}
\end{table}

\begin{table*}[h!]

\setlength{\tabcolsep}{12pt}
\centering
\scalebox{0.85}{
\begin{tabular}{lccccccc}
\toprule
  Models & PATO & WN\textsubscript{animal.n.01}& WN\textsubscript{plant.n.02} &  HDO & HPO & GO& WN\textsubscript{entity.n.01} \\
 \midrule
 BOW-LR &   0.79 & 0.75 & 0.65 & 0.55 & 0.63 & 0.32 & 0.41\\
MS-LSTM\textsubscript{$\lambda=0$} & 0.77 & 0.73 & 0.62 & 0.70 & 0.72 & 0.69 & 0.51 \\
MS-LSTM\textsubscript{$\lambda=0.5$} & 0.80 &  0.76 & 0.65 &  0.70 &  0.73 &  0.70 & 0.57 \\
MS-LSTM\textsubscript{$\lambda=1$} & 0.75 & 0.66 & 0.57 & 0.65 & 0.63 & 0.62 & 0.51 \\

  text2nodes & 0.75 & 0.66 & 0.66 & 0.69 & 0.62 & 0.67 & 0.60 \\
  
  text2edges & 0.76 & 0.68 &  0.66 & 0.69 & 0.69 & 0.69 &  0.61 \\
%\hline
  
MS-LSTM$^*_{\lambda=0.5}$ & 0.81 & 0.76 & 0.66 &  0.71 &0.74 & 0.71 & 0.58 \\
text2nodes$^*$ & 0.83 & 0.71 & 0.68 & 0.71 & 0.69 & 0.70 & 0.62 \\
text2edges$^*$ & \bf 0.83 & \bf 0.77 & \bf 0.70 & \bf 0.73 & \bf 0.74 & \bf 0.72 & \bf 0.65 \\
\bottomrule
\end{tabular}
}
\caption{Ancestor F1 results. Numbers in bold represent the best performing system on a graph. Models marked with $*$ make use of pre-trained word embedding in their encoder. Lambda ($\lambda$) is defined in Section \ref{baseline_sec}.  We use the same number of epochs, batch size and number of latent dimensions both for MS-LSTM and our models (Appendix \ref{C}). } 
  \label{table:main_res}
\end{table*}

\paragraph{Ontology Preprocessing.}
 
All the ontologies we experimented with are represented as directed acyclic graphs (DAGs).
This creates an ambiguity for node path definitions since there are multiple pathways from a root concept to other concepts. We have assumed that a single unambiguous pathway will reduce the complexity of the problem and leave the comparison with ambiguous pathways (which would inevitably involve a more complex model) to future work.
%To reduce the complexity of the problem, in our ongoing work, we simplify the graph to a tree i.e. there is exist unique path between entities in the ontology.
To convert a DAG to a tree we constrain each entity to have only one parent node. The edges between the other parent nodes are removed.\footnote{The choice of an edge is performed on random basis.}

\paragraph{Path Representations.}
We also experiment with two path representations.  Our first approach, \emph{text2nodes}, uses the label of an entity (cf. Section \ref{introduction}) to represent a path. This is not 
efficient since the decoder of the model needs to select between all of the entities in an ontology and also requires more parameters in the model.  Our second approach, \emph{text2edges}, to reduce the number of symbols for the model to choose from, uses edges to represent the path.
To do this we create an artificial vocabulary of the size $\Delta(G)$, where $\Delta(G)$ corresponds to the maximum degree of a node. Each edge in the graph is labeled using the artificial vocabulary. For the example in Section \ref{introduction}, the path would be
\emph{animal} $-$[a]$\rightarrow$ \emph{chordate} $-$[b]$\rightarrow$ \emph{vertebrate} $-$[c]$\rightarrow$ \emph{bird} $-$[d]$\rightarrow$ \emph{apodiform\_bird} where \{a,b,c,d\} is the artificial vocabulary. In the resulting path we discard labels for the entities; therefore, the path reduces to: [a]$\rightarrow$ [b]$\rightarrow$ [c]$\rightarrow$ [d].

\subsection{Baselines}
\label{baseline_sec}
\paragraph*{Bag-of-Words Linear Regression (BOW-LR):} To represent a textual definition in a vector space we first use a pre-trained set of word embeddings \citep{speer2017conceptnet} to represent words in the definition and then find the mean of the word embeddings. As for the ontology, we use node2vec \citep{DBLP:journals/corr/GroverL16}, to represent each entity in a vector space. To align the two vector spaces we use linear regression. %At test time we map a textual definition of a new entity to the ontology vector space and find the closest entity in the graph using the cosine similarity. 
\paragraph*{Multi-Sense LSTM (MS-LSTM):}\citet{Q123} proposed a model that achieves state-of-the-art results on the text-to-entity mapping on the \emph{Snomed CT}\footnote{\url{https://www.snomed.org/snomed-ct}} dataset. The approach uses a novel multi-sense LSTM, augmented with an attention mechanism, to project the definition to the ontology vector space. Additionally, for a better alignment between the two vector spaces, the authors augmented the ontology graph with textual features.

\subsection{Evaluation Metric}
To perform evaluation of the models described above we used Ancestor-F1 score \citep{P18-1229}.  This metric compares the ancestors ($is-a_{model}$)  of the predicted node with the ancestors ($is-a_{gold}$)  of the gold node in the taxonomy.
\[P = \frac{|is-a_{model} \wedge is-a_{gold}|}{|is-a_{model}|},\]
\[R = \frac{|is-a_{model} \wedge is-a_{gold}|}{|is-a_{gold}|},\]
where $P$ and $R$ are precision and recall, respectively. The Ancestor-F1 is then defined as: \[ 2 \times \frac{P \times R}{P + R}. \]

\subsection{Intrinsic Evaluation}
\label{results_sec}

To verify the reliability of our model on text-to-entity mapping we did a set of experiments on the seven graphs (Section \ref{ontologies}) where we map a textual definition of a concept to a path.

To conduct the experiments we randomly sampled 10\% of leaves from the graph.  From this sample, 90\% are used to evaluate the model and 10\% are used to tune the model. The remaining nodes in the graph are used for training. We sample leaves for two reasons: (1) to predict a leaf, the model needs to make the maximum number of (correct) predictions and (2) this way we do not change the original topology of the graph. Note that the sampled nodes and their textual definitions are not present in the training data.

Both baselines predict a single entity instead of a path. To have the same evaluation framework for all the models, for each node predicted by the baselines we create\footnote{We used NetworkX (\url{https://networkx.github.io}) to find a path from predicted node to the root of a graph.} a path from the root of the node to the predicted node. %This way we can use Ancestor-F1 for all the models.
However, we want to emphasize that this is disadvantageous for our model, since all the symbols in the path are predicted by it and in the case of the baselines only a single node is predicted.

The results are presented in Table \ref{table:main_res}. Models that are in the last three rows of Table \ref{table:main_res} use pre-trained word embeddings \citep{speer2017conceptnet} in the encoder. MS-LSTM and our models that are above the last three rows use randomly initialised word vectors. 
We had four observations: (1) without pre-trained word embeddings in the encoder our model outperforms the best MS-LSTM\textsubscript{$\lambda=0.5$} only on two of the seven graphs, (2) the text2edges$^*$ model outperforms all the other models including MS-LSTM$^*_{\lambda=0.5}$, (3) the text2edges model can better exploit pre-trained word embeddings than MS-LSTM, (4) our model performs better when the paths are represented using edges (rather than nodes). We also found that there is a strong negative correlation (Spearman: $-0.75$, Pearson: $-0.80$) between A.D. (Table \ref{ontologies}) and the Ancestor F1 score for the text2edges$^*$ model, meaning that with an increase in A.D. the Ancestor F1 score decreases.

\subsection{Error Analysis}

We carried out an analysis on the outputs of our best-performing model, i.e. text2edges$^*$ with pre-trained word embeddings.
One factor that affects the performance is the number of invalid sequences predicted by the text2nodes and text2edges models. An invalid sequence is the path that does not exist in the original graph. This happens because at each time step the decoder outputs a distribution over all the nodes/edges and not just over possible children nodes. We therefore performed a count of the number of invalid sequences produced by the model.  The percentage of invalid sequences is in the range of 1.82\% - 8.50\% (Appendix \ref{B}), which is relatively low. This analysis was also performed by \citet{GrammarVAE}. To guarantee that the model always produces valid graphs, they use a context-free grammar. A similar method can be adapted in our work.

\begin{figure}[H]
\centering
\begin{tabular}{@{}c@{}}
 
  \includegraphics[trim={0.5cm 0cm 0 0.5cm}, width=0.96\linewidth]{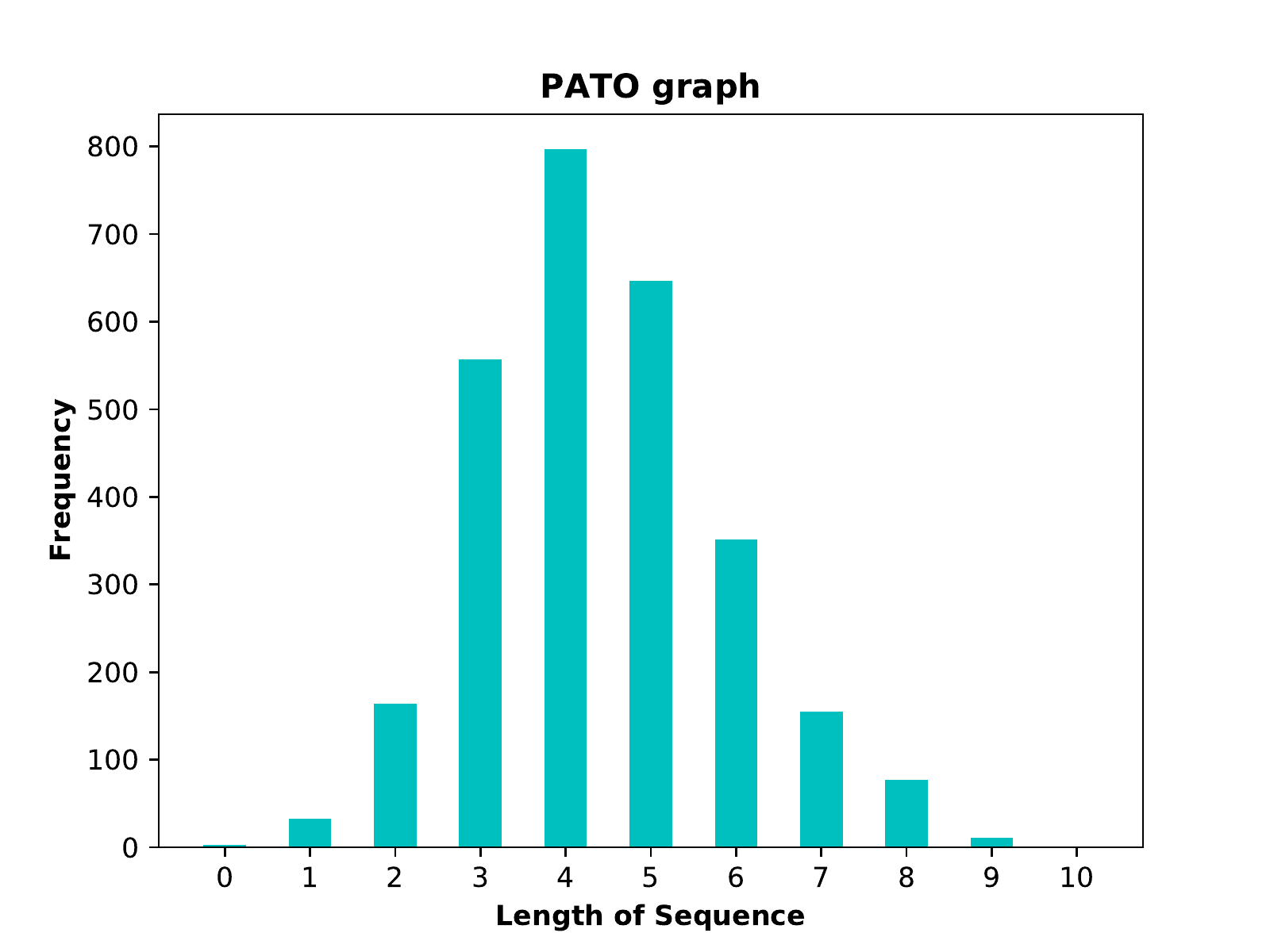}

\end{tabular}%
\vspace{-3.1mm}
\begin{tabular}{@{}c@{}}

  \includegraphics[width=0.96\linewidth, trim={0.5cm 0.2cm 0 0cm}]{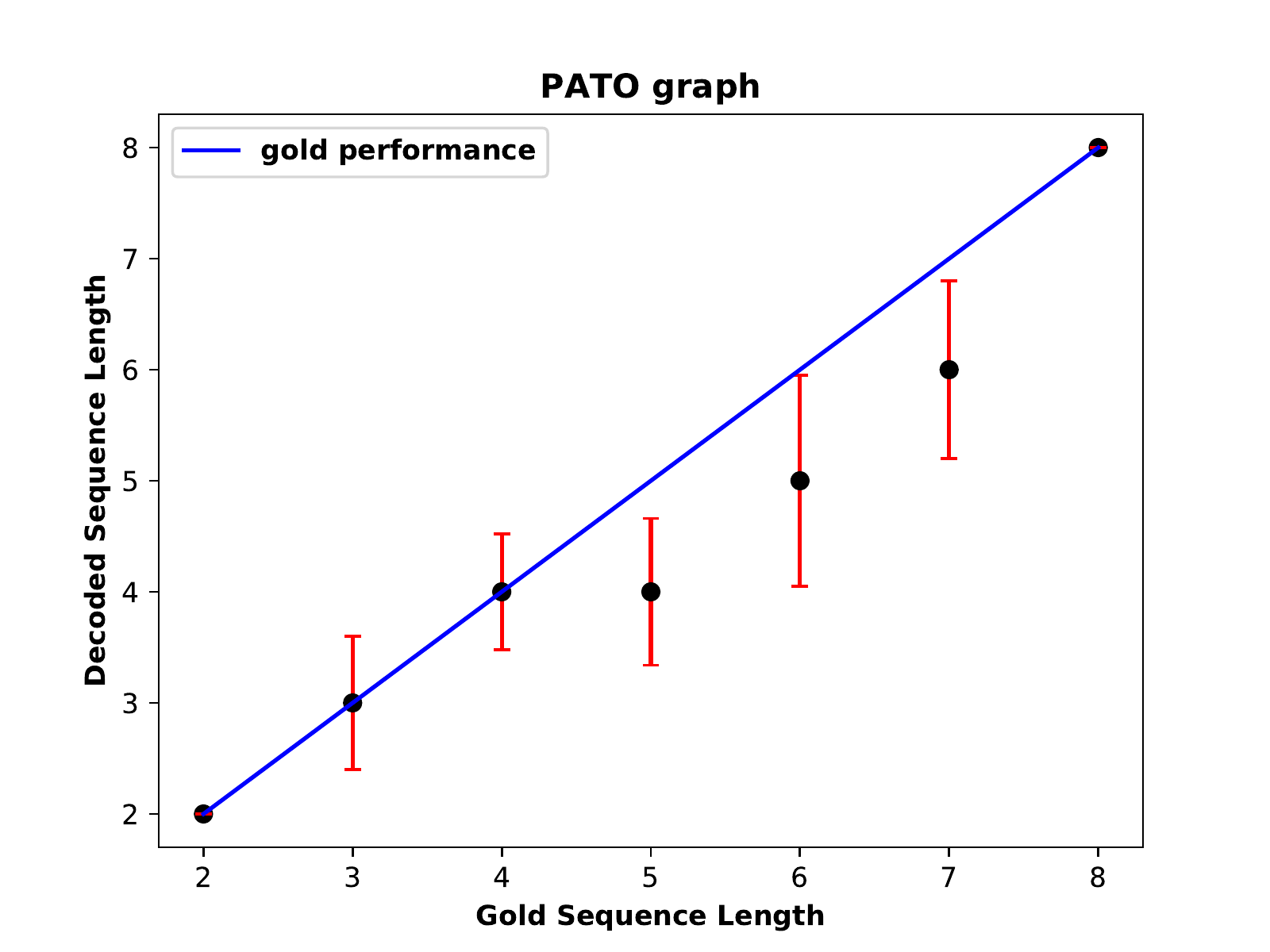}
  
\end{tabular}
\caption{The graph on top shows the length of 
sequence vs length frequency on a training set. The graph on the bottom shows the length of the gold sequence vs mean length of decoded sequence on the test set.}
\label{fig:stats}
\end{figure}

Another factor that affects the performance is the length of the generated paths which is expected to match the length of the gold path. To test this, we compared the mean length of the generated sequences with the length of the gold path (the graph on the bottom of Figure \ref{fig:stats}). Also, in the training set, we associate the length of the sequences with their frequencies (the graph on the top of Figure \ref{fig:stats}).
We found that (1) the length of the generated paths are biased towards the more frequent paths in the training data, (2) if the length of a path is not frequent in the training data, the model either under-generates or over-generates the length (Appendix \ref{D}).  

\section{Related Work}
Text-to-entity mapping is an essential component of many NLP tasks, e.g. fact verification \citep{N18-1074} or question answering \citep{Yih2015SemanticPV}. Previous work has approached this problem with pairwise learning-to-rank method \cite{a123} or phrase-based machine translation \cite{DBLP:journals/corr/LimsopathamC15}. However, these methods generally ignore ontology's structure.
%{\bf \color{red}Write one/two sentences why these techniques are not good.}
More recent work has viewed the problem of text-to-entity mapping as a projection of a textual definition to a single point in a KG \citep{Q123, DBLP:journals/corr/HillCKB15}. However, despite potential advantages, such as being more interpretable and  less brittle (model predicts multiple related entities instead of one), path-based approaches have received relatively little attention. Instead of predicting a single entity, path-based models, such as the one we proposed in this paper, try to map a textual definition to multiple relevant entities in an external resource. 

\section{Conclusion and Future Work}
We presented a model that maps textual definitions to interpretable ontological pathways. We evaluated the proposed technique on seven semantic graphs, showing that it can perform competitively with respect to existing state-of-the-art text-to-entity systems, while being more interpretable and self-contained. 
We hope this work will encourage further research on path-based text-to-entity mapping algorithms.
A natural next step will be to extend our framework to DAGs. Furthermore, we plan to constrain our model to always predict paths that exist in the graph, as we discussed above. 

\section*{Acknowledgments}

We would like to thank the anonymous reviewers for their comments. Also, we would like to thank Dimitri Kartsaklis and Ehsan Shareghi for helpful discussions and comments. This research was supported by an EPSRC Experienced Researcher Fellowship (N. Collier: EP/M005089/1) and an MRC grant (M.T. Pilehvar: MR/M025160/1). We gratefully acknowledge the donation of a GPU from the NVIDIA Grant Program.

\bibliography{naaclhlt2019}

\begin{thebibliography}{18}
\expandafter\ifx\csname natexlab\endcsname\relax\def\natexlab#1{#1}\fi

\bibitem[{Ashburner et~al.(2000)Ashburner, Ball, Blake, Botstein, Butler,
  Cherry, Davis, Dolinski, Dwight, Eppig, Harris, Hill, Issel-Tarver,
  Kasarskis, Lewis, Matese, Richardson, Ringwald, Rubin, and
  Sherlock}]{Ashburner_2000}
Michael Ashburner, Catherine~A. Ball, Judith~A. Blake, David Botstein, Heather
  Butler, J.~Michael Cherry, Allan~P. Davis, Kara Dolinski, Selina~S. Dwight,
  Janan~T. Eppig, Midori~A. Harris, David~P. Hill, Laurie Issel-Tarver, Andrew
  Kasarskis, Suzanna Lewis, John~C. Matese, Joel~E. Richardson, Martin
  Ringwald, Gerald~M. Rubin, and Gavin Sherlock. 2000.
\newblock \href {https://doi.org/10.1038/75556} {Gene ontology: tool for the
  unification of biology}.
\newblock \emph{Nature Genetics}, 25(1):25--29.

\bibitem[{Bahdanau et~al.(2014)Bahdanau, Cho, and
  Bengio}]{DBLP:journals/corr/BahdanauCB14}
Dzmitry Bahdanau, Kyunghyun Cho, and Yoshua Bengio. 2014.
\newblock \href {http://arxiv.org/abs/1409.0473} {Neural machine translation by
  jointly learning to align and translate}.
\newblock \emph{CoRR}, abs/1409.0473.

\bibitem[{Fellbaum(1998)}]{Fellbaum:98}
Christiane Fellbaum, editor. 1998.
\newblock \emph{{W}ord{N}et: An Electronic Database}.
\newblock MIT Press, Cambridge, MA.

\bibitem[{Grover and Leskovec(2016)}]{DBLP:journals/corr/GroverL16}
Aditya Grover and Jure Leskovec. 2016.
\newblock \href {http://arxiv.org/abs/1607.00653} {node2vec: Scalable feature
  learning for networks}.
\newblock \emph{CoRR}, abs/1607.00653.

\bibitem[{Hill et~al.(2015)Hill, Cho, Korhonen, and
  Bengio}]{DBLP:journals/corr/HillCKB15}
Felix Hill, Kyunghyun Cho, Anna Korhonen, and Yoshua Bengio. 2015.
\newblock \href {http://arxiv.org/abs/1504.00548} {Learning to understand
  phrases by embedding the dictionary}.
\newblock \emph{CoRR}, abs/1504.00548.

\bibitem[{Hochreiter and
  Schmidhuber(1997)}]{Hochreiter:1997:LSM:1246443.1246450}
Sepp Hochreiter and J\"{u}rgen Schmidhuber. 1997.
\newblock \href {https://doi.org/10.1162/neco.1997.9.8.1735} {Long short-term
  memory}.
\newblock \emph{Neural Comput.}, 9(8):1735--1780.

\bibitem[{J.~Kusner et~al.(2017)J.~Kusner, Paige, and Miguel
  HernÃ¡ndez-Lobato}]{GrammarVAE}
Matt J.~Kusner, Brooks Paige, and JosÃ© Miguel HernÃ¡ndez-Lobato. 2017.
\newblock Grammar variational autoencoder.

\bibitem[{Kartsaklis et~al.(2018)Kartsaklis, Pilehvar, and Collier}]{Q123}
Dimitri Kartsaklis, Mohammad~Taher Pilehvar, and Nigel Collier. 2018.
\newblock \href {http://aclweb.org/anthology/D18-1221} {Mapping text to
  knowledge graph entities using multi-sense lstms}.
\newblock In \emph{Proceedings of the 2018 Conference on Empirical Methods in
  Natural Language Processing}, pages 1959--1970. Association for Computational
  Linguistics.

\bibitem[{Leaman et~al.(2013)Leaman, Dogan, and lu}]{a123}
Robert Leaman, Rezarta Dogan, and Zhiyong lu. 2013.
\newblock \href {https://doi.org/10.1093/bioinformatics/btt474} {Dnorm: Disease
  name normalization with pairwise learning to rank}.
\newblock \emph{Bioinformatics (Oxford, England)}, 29.

\bibitem[{Limsopatham and Collier(2015)}]{DBLP:journals/corr/LimsopathamC15}
Nut Limsopatham and Nigel Collier. 2015.
\newblock \href {http://arxiv.org/abs/1508.02285} {Adapting phrase-based
  machine translation to normalise medical terms in social media messages}.
\newblock \emph{CoRR}, abs/1508.02285.

\bibitem[{Mao et~al.(2018)Mao, Ren, Shen, Gu, and Han}]{P18-1229}
Yuning Mao, Xiang Ren, Jiaming Shen, Xiaotao Gu, and Jiawei Han. 2018.
\newblock \href {http://aclweb.org/anthology/P18-1229} {End-to-end
  reinforcement learning for automatic taxonomy induction}.
\newblock In \emph{Proceedings of the 56th Annual Meeting of the Association
  for Computational Linguistics (Volume 1: Long Papers)}, pages 2462--2472.
  Association for Computational Linguistics.

\bibitem[{Robinson et~al.(2008)Robinson, K{\"o}hler, Bauer, Seelow, Horn, and
  Mundlos}]{Robinson2008TheHP}
Peter~N. Robinson, Sebastian K{\"o}hler, Sebastian~B Bauer, Dominik Seelow,
  Denise Horn, and Stefan Mundlos. 2008.
\newblock The human phenotype ontology: a tool for annotating and analyzing
  human hereditary disease.
\newblock \emph{American journal of human genetics}, 83 5:610--5.

\bibitem[{Schriml et~al.(2012)Schriml, Arze, Nadendla, Chang, Mazaitis, Felix,
  Feng, and Kibbe}]{Schriml2012DiseaseOA}
Lynn~M. Schriml, Cesar Arze, Suvarna Nadendla, Yu-Wei~Wayne Chang, Mark
  Mazaitis, Victor Felix, Gang Feng, and Warren~A. Kibbe. 2012.
\newblock Disease ontology: a backbone for disease semantic integration.
\newblock In \emph{Nucleic Acids Research}.

\bibitem[{Sennrich et~al.(2016)Sennrich, Haddow, and Birch}]{P16-1009}
Rico Sennrich, Barry Haddow, and Alexandra Birch. 2016.
\newblock \href {https://doi.org/10.18653/v1/P16-1009} {Improving neural
  machine translation models with monolingual data}.
\newblock In \emph{Proceedings of the 54th Annual Meeting of the Association
  for Computational Linguistics (Volume 1: Long Papers)}, pages 86--96.
  Association for Computational Linguistics.

\bibitem[{Speer et~al.(2017)Speer, Chin, and Havasi}]{speer2017conceptnet}
Robert Speer, Joshua Chin, and Catherine Havasi. 2017.
\newblock \href {http://aaai.org/ocs/index.php/AAAI/AAAI17/paper/view/14972}
  {Conceptnet 5.5: An open multilingual graph of general knowledge}.

\bibitem[{Sutskever et~al.(2014)Sutskever, Vinyals, and
  Le}]{DBLP:journals/corr/SutskeverVL14}
Ilya Sutskever, Oriol Vinyals, and Quoc~V. Le. 2014.
\newblock \href {http://arxiv.org/abs/1409.3215} {Sequence to sequence learning
  with neural networks}.
\newblock \emph{CoRR}, abs/1409.3215.

\bibitem[{Thorne et~al.(2018)Thorne, Vlachos, Christodoulopoulos, and
  Mittal}]{N18-1074}
James Thorne, Andreas Vlachos, Christos Christodoulopoulos, and Arpit Mittal.
  2018.
\newblock \href {https://doi.org/10.18653/v1/N18-1074} {Fever: a large-scale
  dataset for fact extraction and verification}.
\newblock In \emph{Proceedings of the 2018 Conference of the North American
  Chapter of the Association for Computational Linguistics: Human Language
  Technologies, Volume 1 (Long Papers)}, pages 809--819. Association for
  Computational Linguistics.

\bibitem[{Yih et~al.(2015)Yih, Chang, He, and Gao}]{Yih2015SemanticPV}
Wen-Tau Yih, Ming-Wei Chang, Xiaodong He, and Jianfeng Gao. 2015.
\newblock Semantic parsing via staged query graph generation: Question
  answering with knowledge base.
\newblock In \emph{ACL}.

\end{thebibliography}
\bibliographystyle{acl_natbib}

\appendix
\section{DAGs}
\label{A}
\begin{table}[H]
\begin{center}
\setlength{\tabcolsep}{12.8pt}
\begin{tabular}{lcc}
 \bf Graphs &  \bf Mult.P\textsubscript{\%} &\bf AV.P \\ 
 \hline
PATO & 31.29 & 2.97\\
WN\textsubscript{animal.n.01} & 0.88 & 2.00\\
WN\textsubscript{plant.n.02} & 0.16 & 2.00\\
HDO & 16.23 & 2.13\\
HPO & 23.24 & 2.23\\
GO  & 64.01 & 2.77\\
WN\textsubscript{entity.n.01} & 1.91 & 2.03 \\
\hline
\end{tabular}
\end{center}
\caption{Statistics of nodes with multiple inheritance. \textbf {Mult.P\textsubscript{\%}} stands for percentage of nodes with more than one parent node. \textbf {AV.P} stands for average number of parents a node with multiple inheritance has. }
\label{table:invalid_seq}
\end{table}

\section{Invalid Sequences}
\label{B}
\begin{table}[H]
\begin{center}
\setlength{\tabcolsep}{13.8pt}
\begin{tabular}{lcc}
 \bf Graphs &  \bf Invalid\textsubscript{\%} &\bf N\textsubscript{total} \\ 
 \hline
PATO & 1.82 & 110\\
WN\textsubscript{animal.n.01} & 4.56 & 263\\
WN\textsubscript{plant.n.02} & 2.23 & 314\\
HDO & 4.02 & 622\\
HPO & 7.08 & 847\\
GO  & 6.94 & 1845\\
WN\textsubscript{entity.n.01} & 8.50 & 5191 \\
\hline

\end{tabular}
\end{center}
\caption{Statistics of invalid sequences. \textbf{Invalid\textsubscript{\%}} is the percentage of invalid sequences and \textbf{N\textsubscript{total}} is the total number of sequences that were tested.}
\label{table:invalid_seq}
\end{table}
\section{Settings for Models}
\label{C}
\paragraph*{BOW-LR:} To represent an ontology in a vector space we use node2vec \url{https://snap.stanford.edu/node2vec/}. For all the graphs the following hyper-parameters of the algorithm are the same: \emph{walk-length= 5}, \emph{window-size=5} and \emph{iter=40}. As for the number of dimensions we set it to 128 for PATO, WN\textsubscript{animal.n.01}, WN\textsubscript{plant.n.02}, HDO and HPO graphs. For GO and WN\textsubscript{entity.n.01} graphs we set it to 256. All the other parameters of node2vec are default.

We do not modify the numberbatch embeddings \url{https://github.com/commonsense/conceptnet-numberbatch}. If a word in a textual definition is missing we initilised the embedding for this word with zeros.

For all the graphs to map the textual vector space into an ontology vector space we use the linear regression model from the \emph{scikit-learn} API \url{https://scikit-learn.org/stable/modules/generated/sklearn.linear_model.LinearRegression.html}
\paragraph*{MS-LSTM:} There are only two hyper-parameters that we vary during the embedding of ontology concepts: \emph{$\lambda$} (we report the values in the paper) and the embedding size of the concepts. We set it to 128 for PATO, WN\textsubscript{animal.n.01}, WN\textsubscript{plant.n.02}, HDO and HPO graphs. For GO and WN\textsubscript{entity.n.01} graphs we set it to 256.

For all the graphs the model is trained for 300 epochs, dimensions of word embeddings is set to 64 and bi-LSTM is used instead of LSTM. Batch size is set to 16 and the number of latent dimensions in bi-LSTM is set to 128 for the PATO, WN\textsubscript{animal.n.01}, WN\textsubscript{plant.n.02}, HDO and HPO graphs. For GO and WN\textsubscript{entity.n.01} graphs we set these parameters to 128 and 256 respectively. All the other hyper-parameters are default.

When we use pre-trained word embeddings we reduce (with PCA \url{https://scikit-learn.org/stable/modules/generated/sklearn.decomposition.PCA.html}) its dimensions from 300 to 64. 

\paragraph{Our Model:} For all the graphs the model is trained for 300 epochs, dimensions of word embeddings (also for node/edges embeddings) is set to 64 and bi-LSTM is used in the encoder and LSTM in the decoder. Batch size is set to 16 and the number of latent dimensions in bi-LSTM encoder and LSTM decoder is set to 128 for the PATO, WN\textsubscript{animal.n.01}, WN\textsubscript{plant.n.02}, HDO and HPO graphs. For GO and WN\textsubscript{entity.n.01} graphs we set these parameters to 128 and 256 respectively. For optimizer we used \emph{RMSProp} (\url{https://www.tensorflow.org/api_docs/python/tf/train/RMSPropOptimizer}) with learning rate = 0.001.

When we use pre-trained word embeddings we reduce (with PCA \url{https://scikit-learn.org/stable/modules/generated/sklearn.decomposition.PCA.html}) its dimensions from 300 to 64. 

\section{Length of Generated Path}
\label{D}
 In Figure 1 and 2 the blue line indicates the ideal scenario i.e. mean length of the generated sequences is equal to the gold length. Black dot is the mean of the length of decoded sequences and the red bars are the standard deviation. One can notice that the general trend is following: for short sequences the mode generates (slightly) longer sequences and for the long sequences it generated (slightly) shorter sequences than the gold standard. Another trend is that the sequences of the certain length are matching the gold standard. To understand why this is happening one needs to look at the graph which relate the length of the sequence in the training corpus and the frequency of this length in the corpus. It is become clear there is a correlation between the two. Such as the model tends to generate the sequence of the length that is presented the most in the training data.
\begin{figure*}[t!]%
\centering
\begin{tabular}{@{}c@{}}
   \includegraphics[height=2in]{pato_gold_vs_decoded_.pdf}%
\end{tabular}
\begin{tabular}{@{}c@{}}
    \includegraphics[height=2in]{pato_gold_vs_decoded.pdf}
\end{tabular}
\vspace{1mm}
\begin{tabular}{@{}c@{}}
    \includegraphics[height=2in]{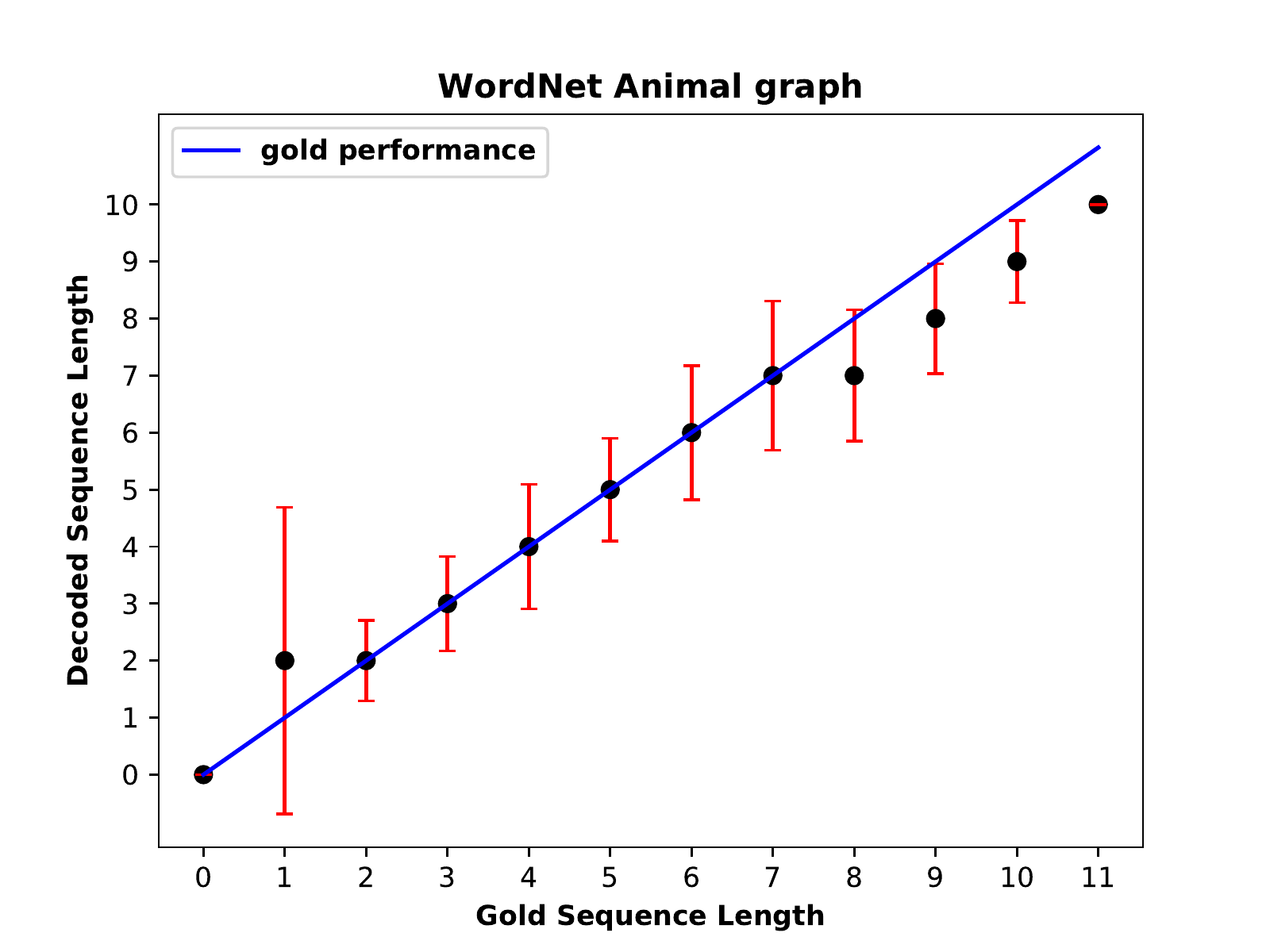}
\end{tabular}
\begin{tabular}{@{}c@{}}
    \includegraphics[height=2in]{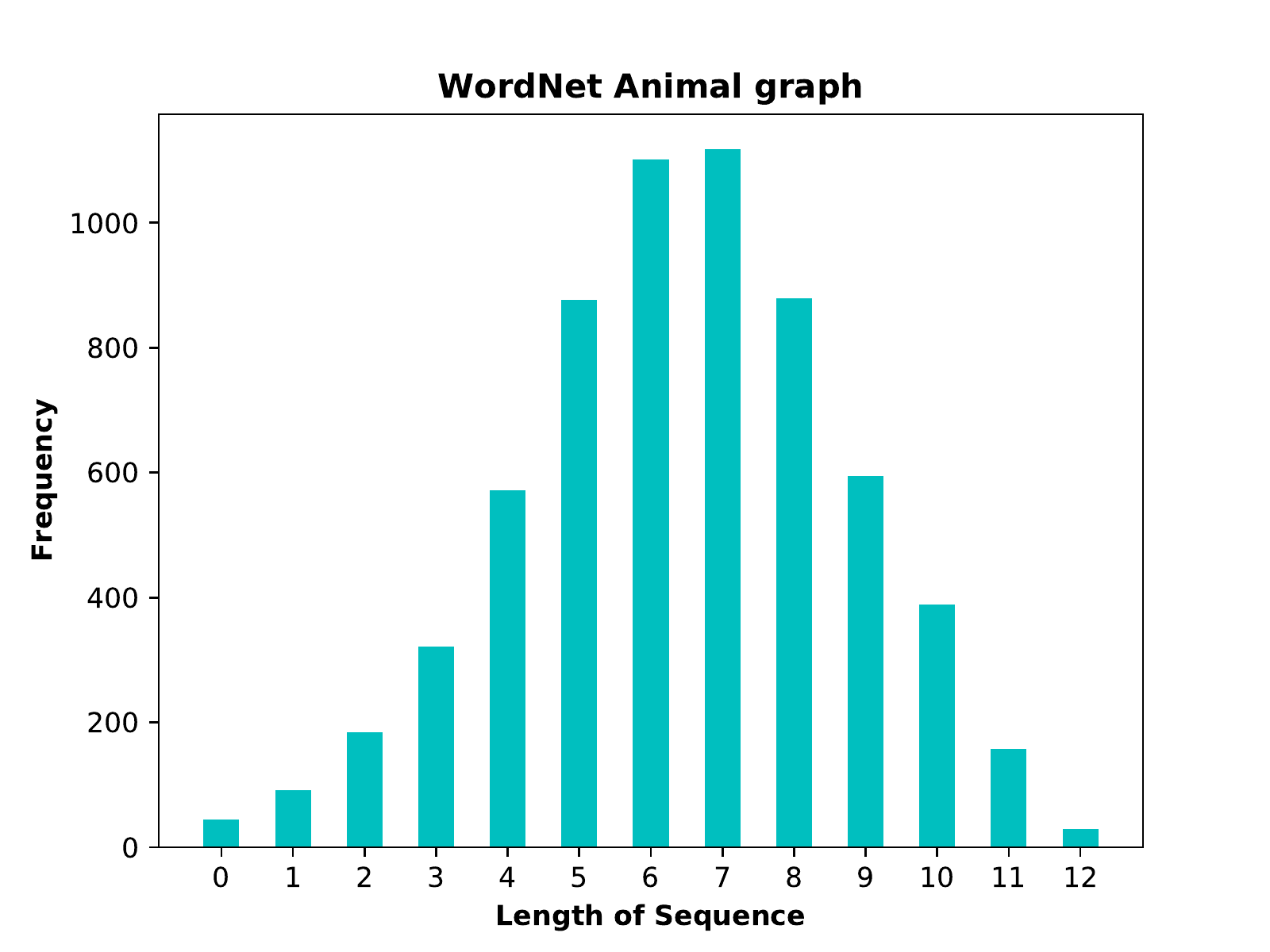}
\end{tabular}
\vspace{1mm}
\begin{tabular}{@{}c@{}}
    \includegraphics[height=2in]{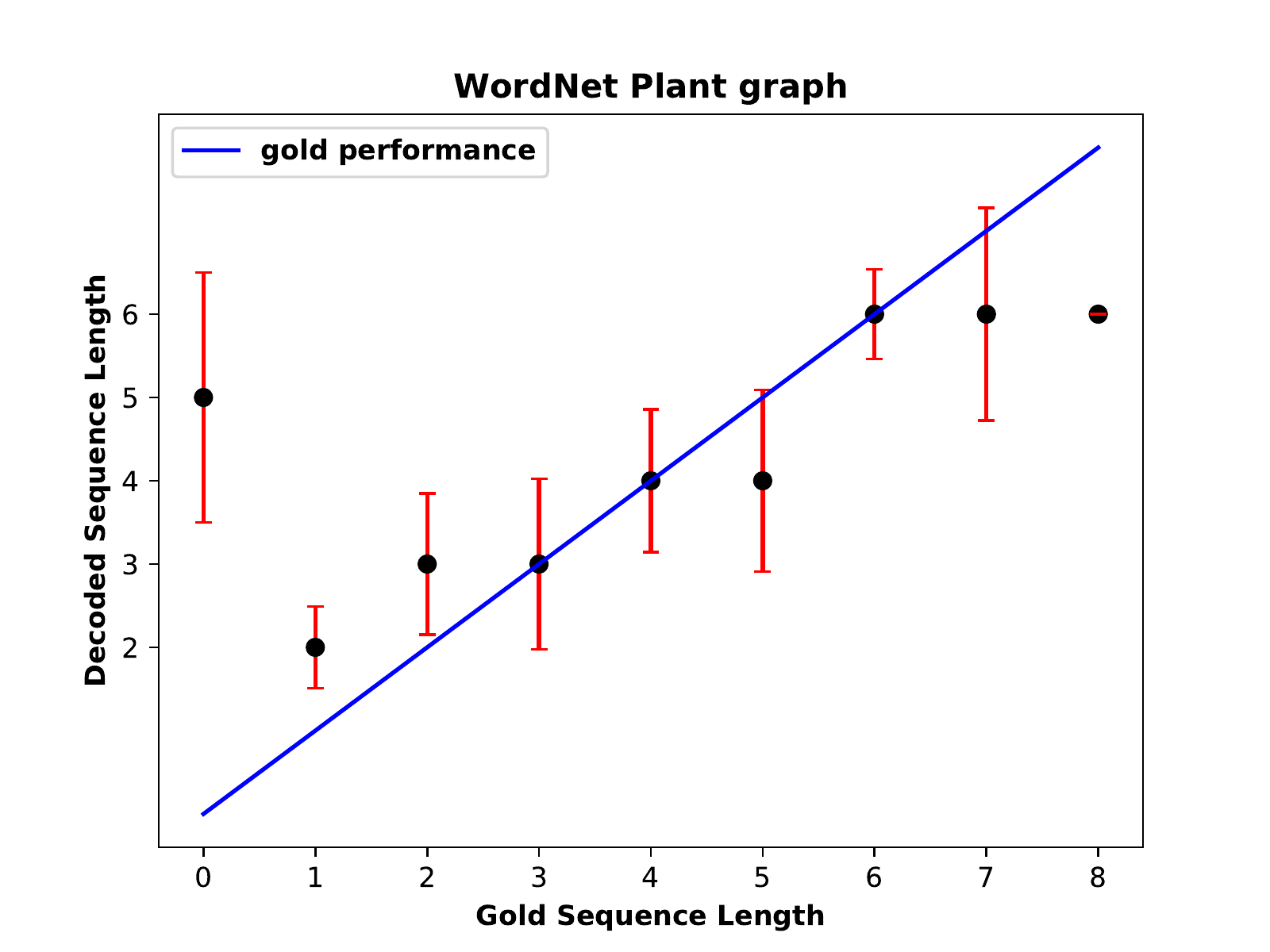}
\end{tabular}
 \begin{tabular}{@{}c@{}}
    \includegraphics[height=2in]{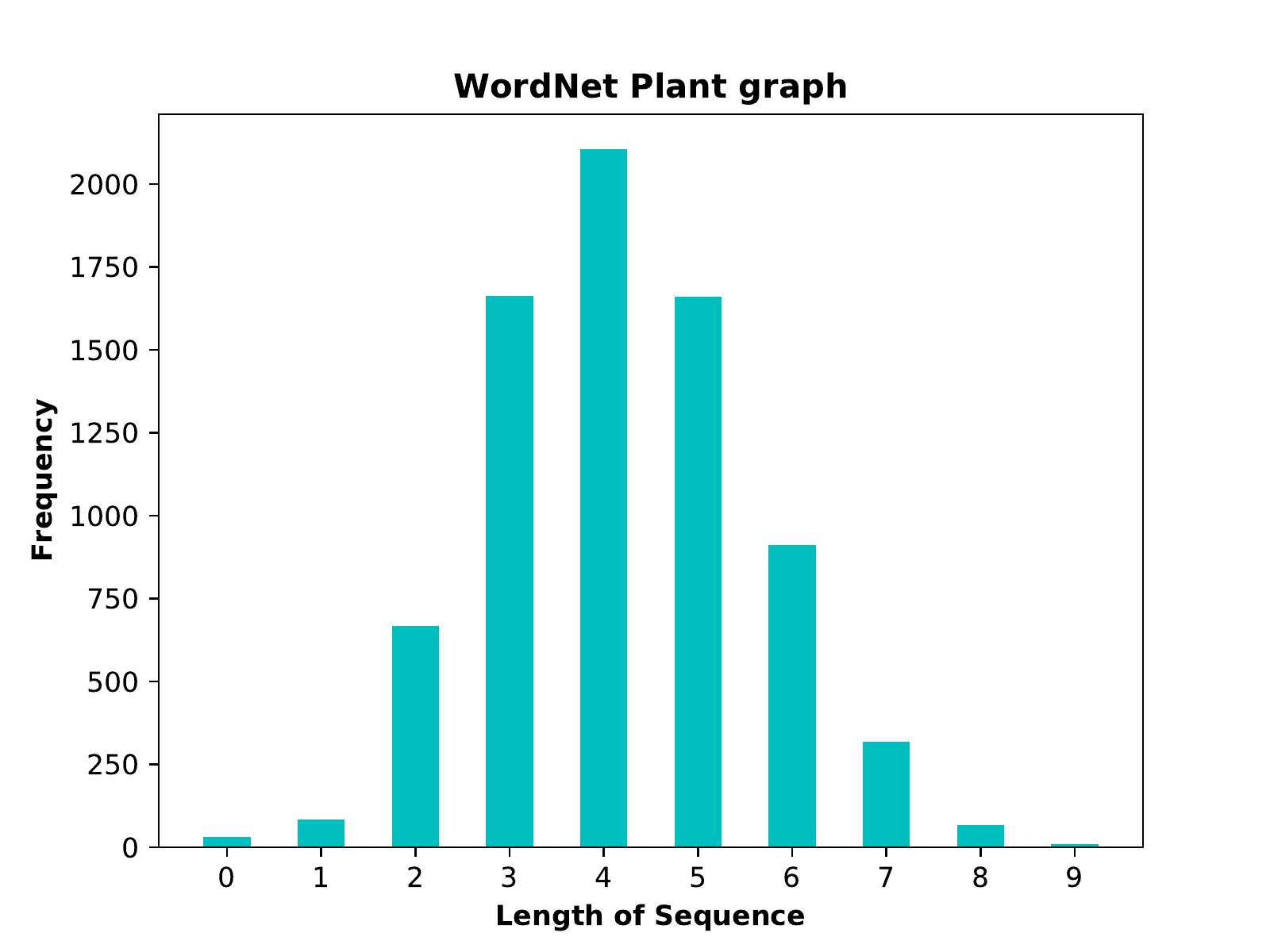}
\end{tabular}
\vspace{1mm}
\begin{tabular}{@{}c@{}}
    \includegraphics[height=2in]{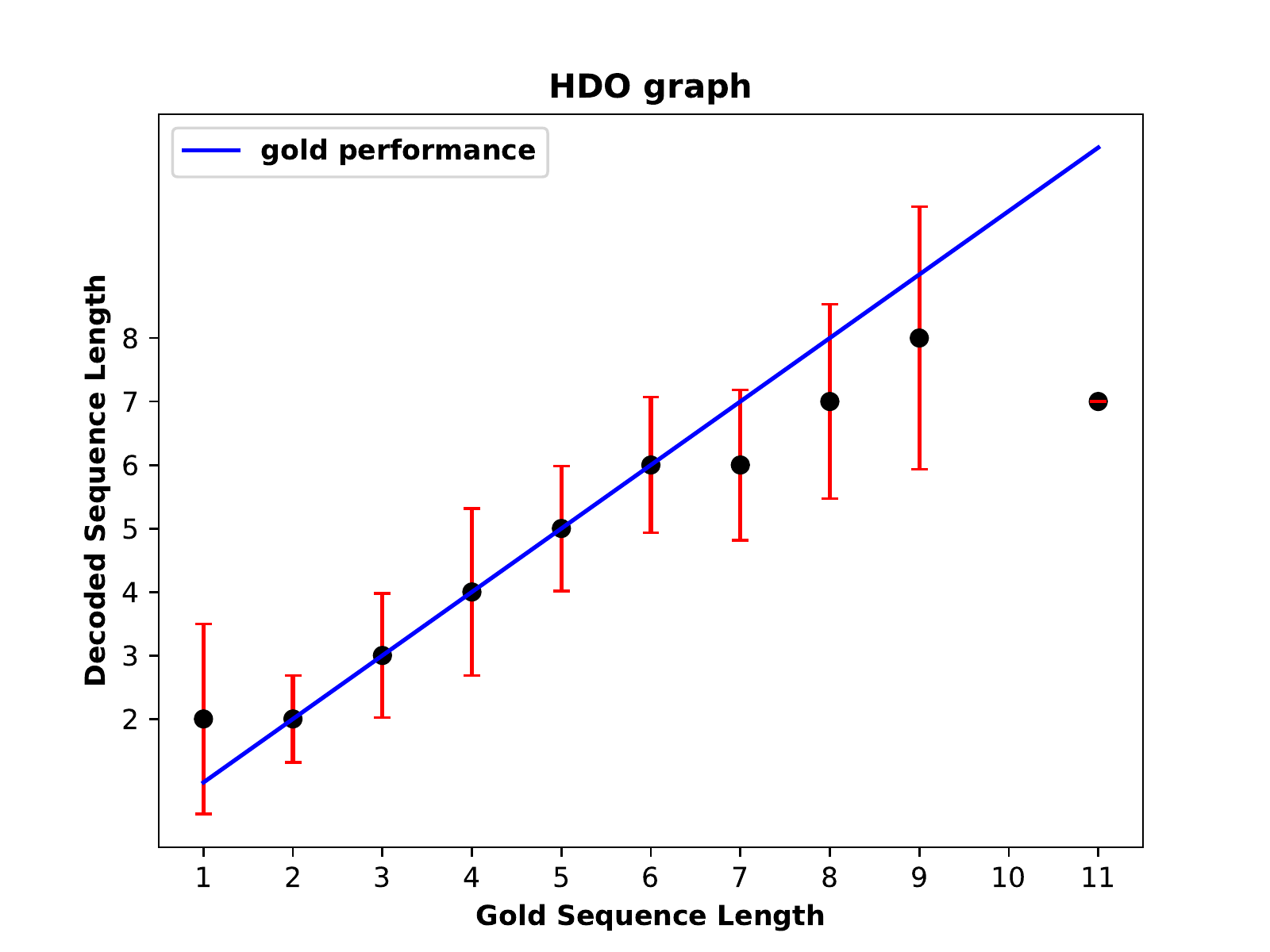}
\end{tabular}
\begin{tabular}{@{}c@{}}
    \includegraphics[height=2in]{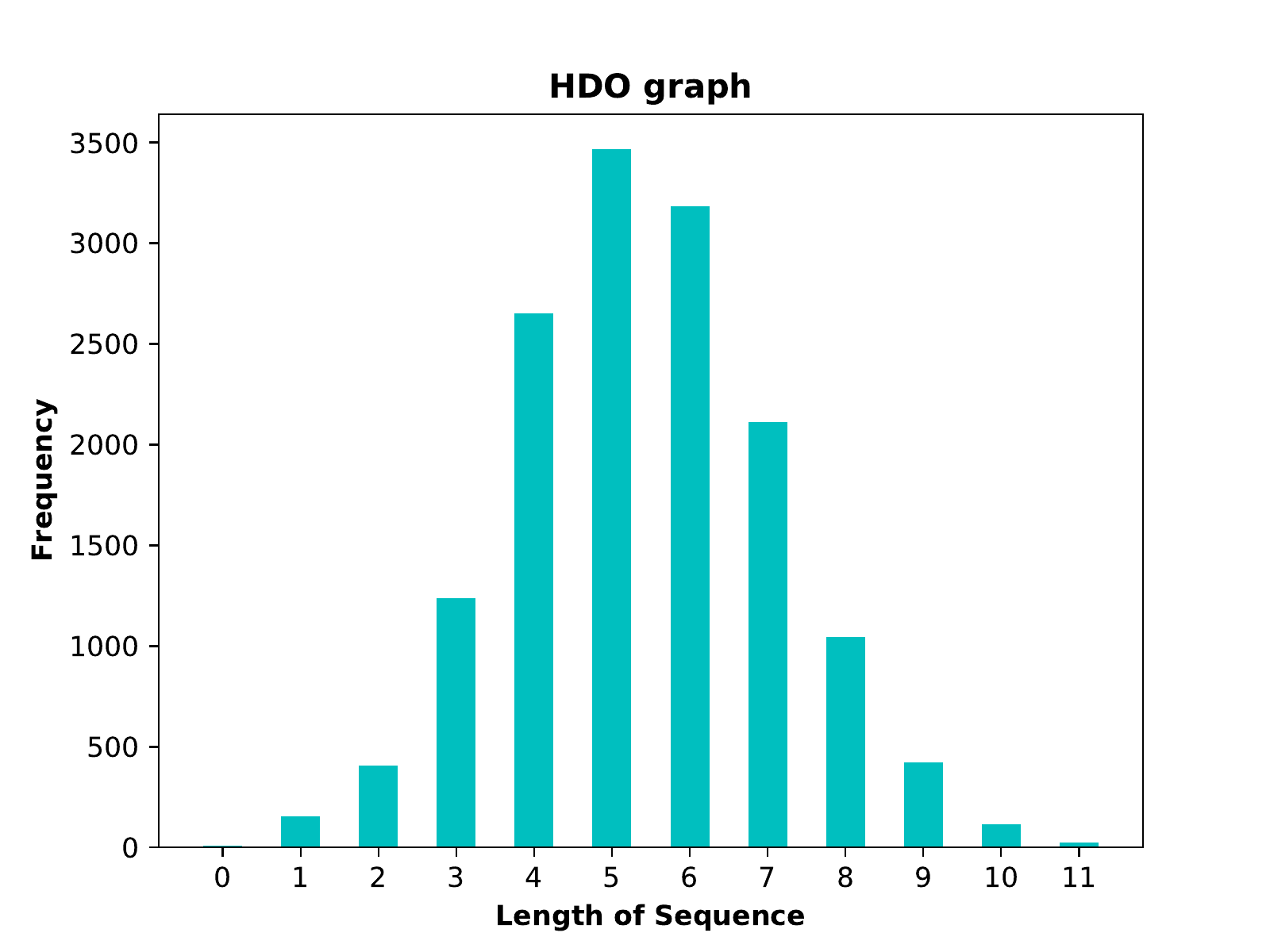}
\end{tabular}

\caption{On the left graphs show: length of gold sequence vs mean length of decoded sequence on a test set; On the right graphs show: length of 
sequence vs length frequency on a training set.}%
%    \end{adjustwidth}
    \label{fig:ex3}%
\end{figure*}

\begin{figure*}[t]%
    \centering
\begin{tabular}{@{}c@{}}
    \includegraphics[height=2in]{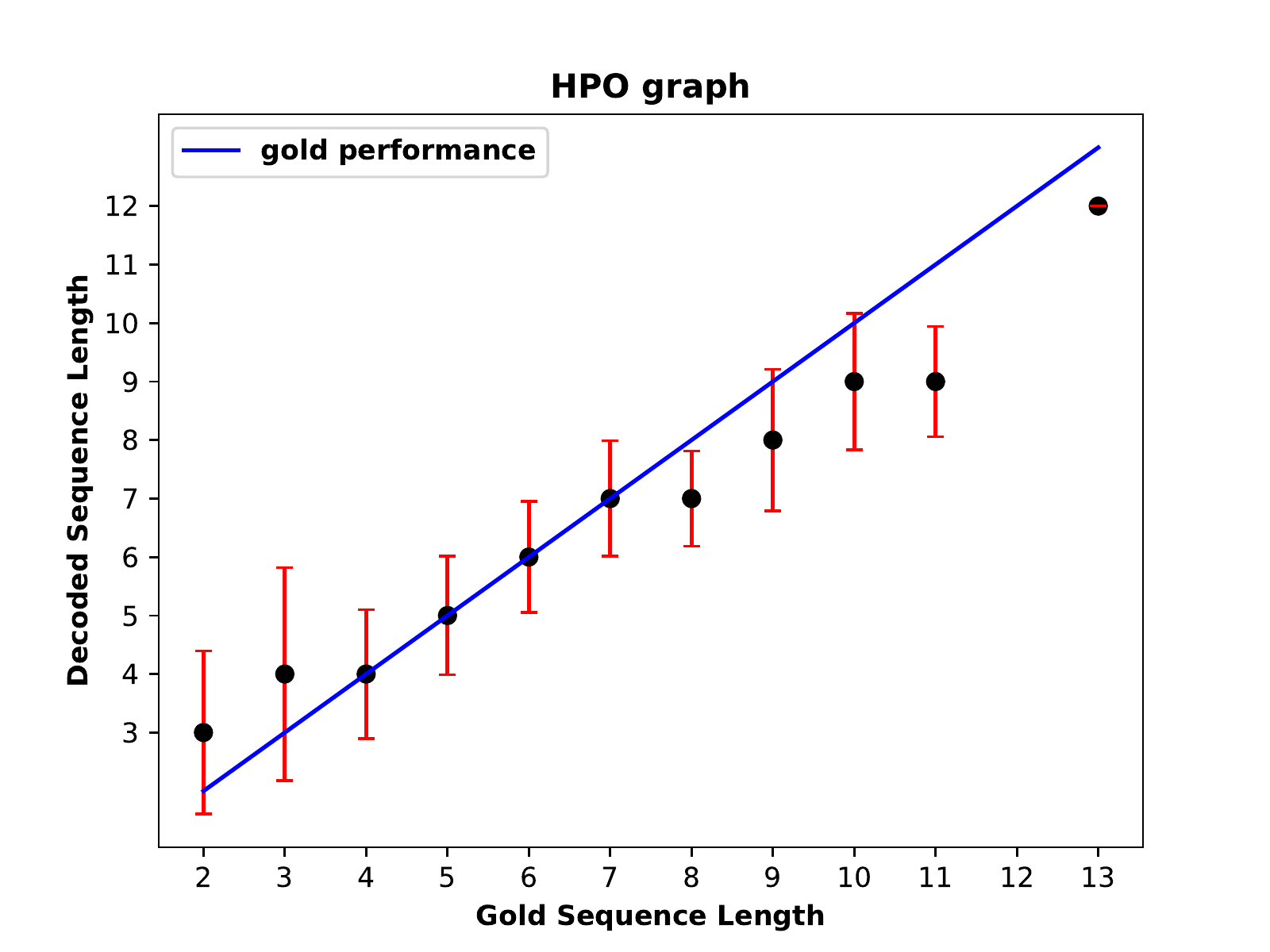}
\end{tabular}
 \begin{tabular}{@{}c@{}}
    \includegraphics[height=2in]{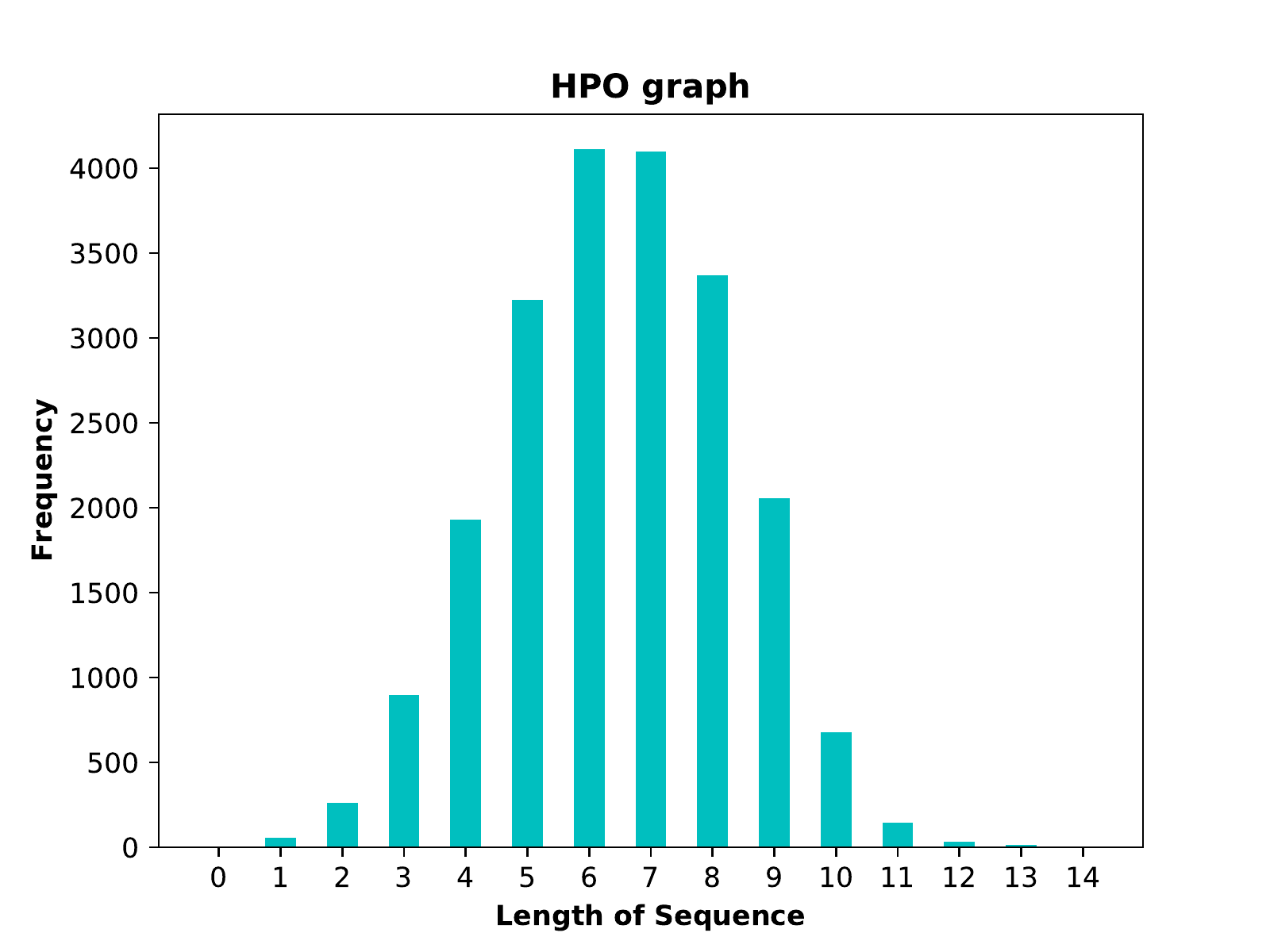}
 \end{tabular}
 \vspace{1mm}
 \begin{tabular}{@{}c@{}}
    \includegraphics[height=2in]{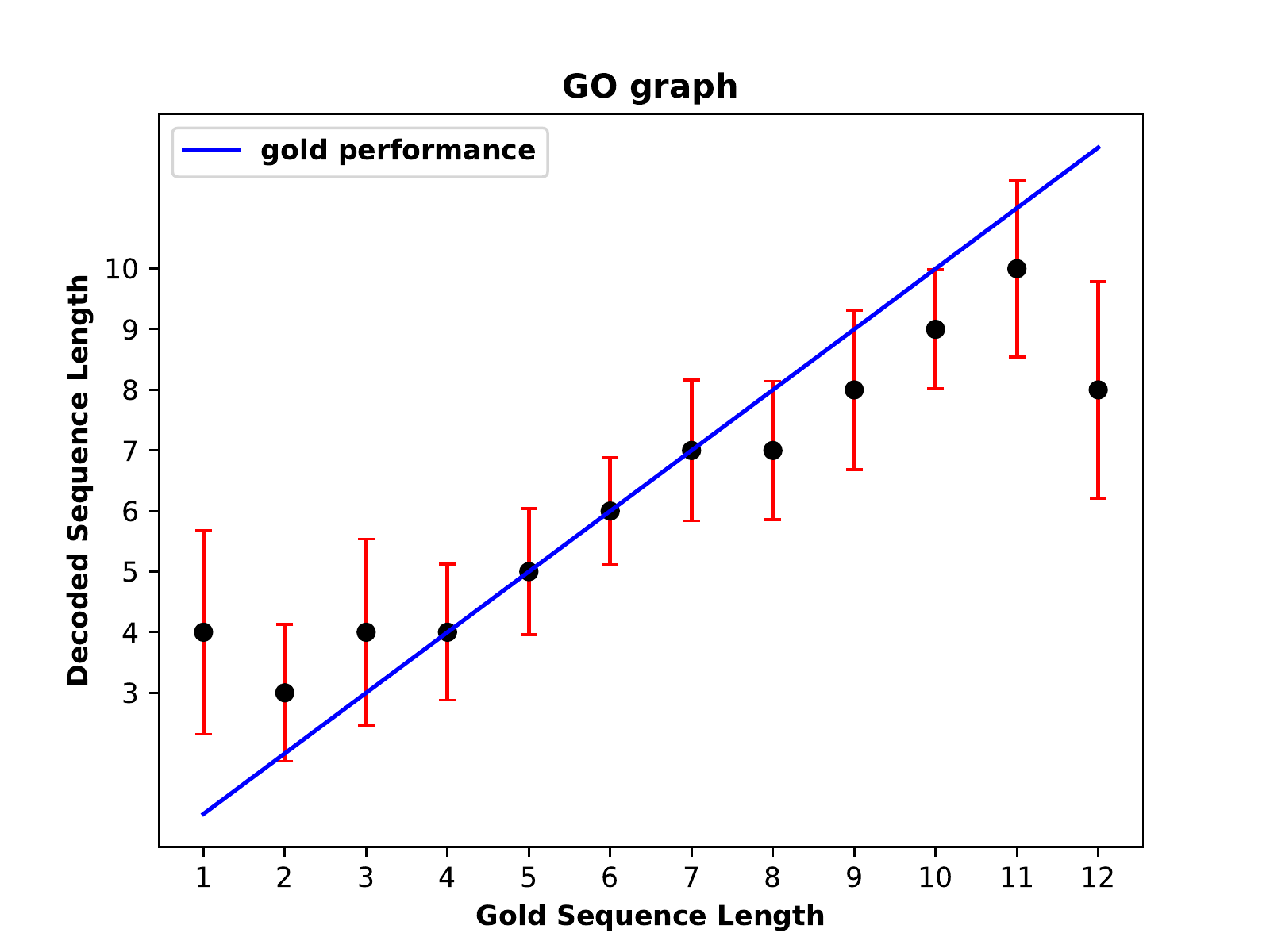}
\end{tabular}
 \begin{tabular}{@{}c@{}}
    \includegraphics[height=2in]{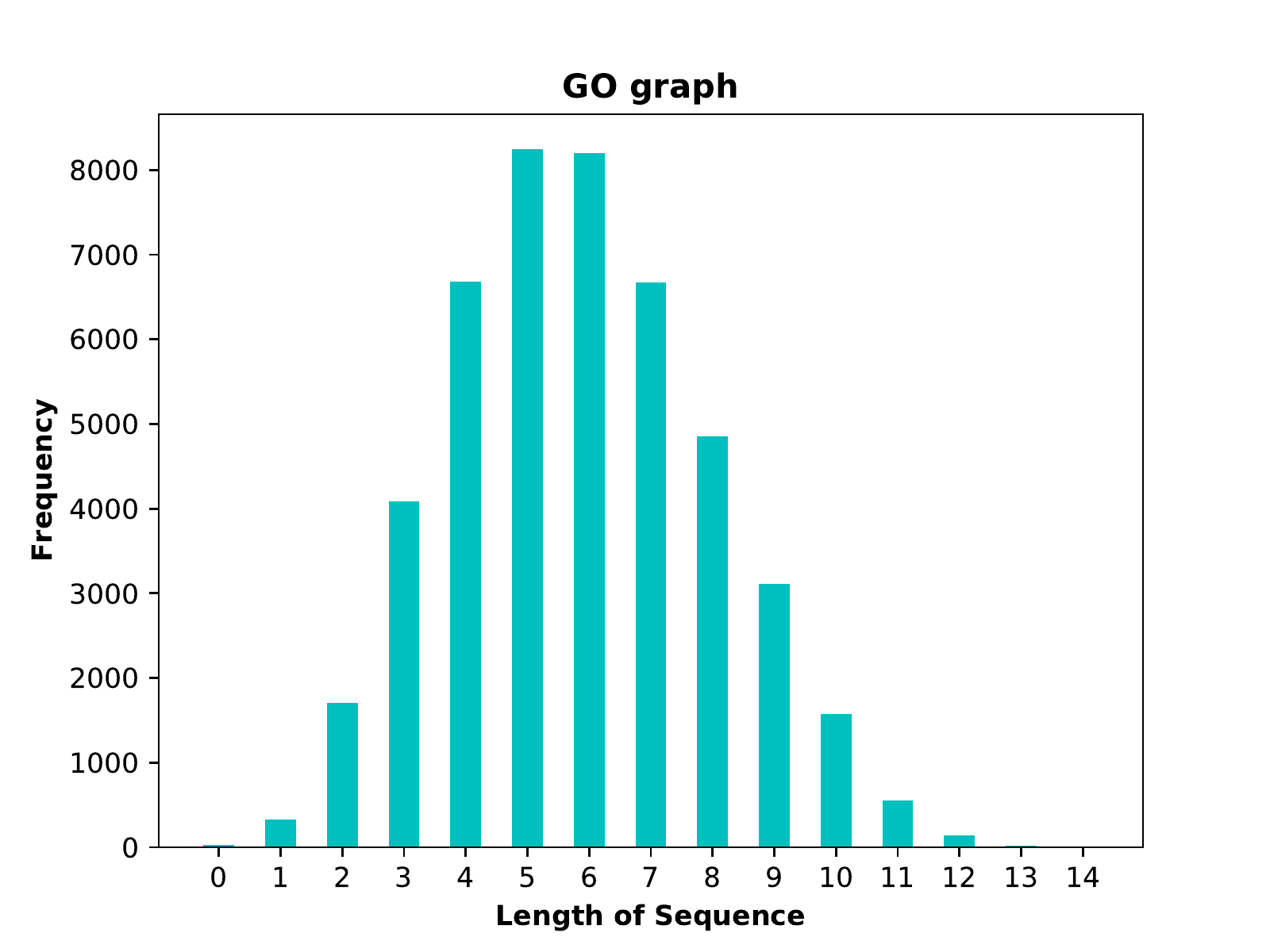}
\end{tabular}
\vspace{1mm}
 \begin{tabular}{@{}c@{}}
    \includegraphics[height=2in]{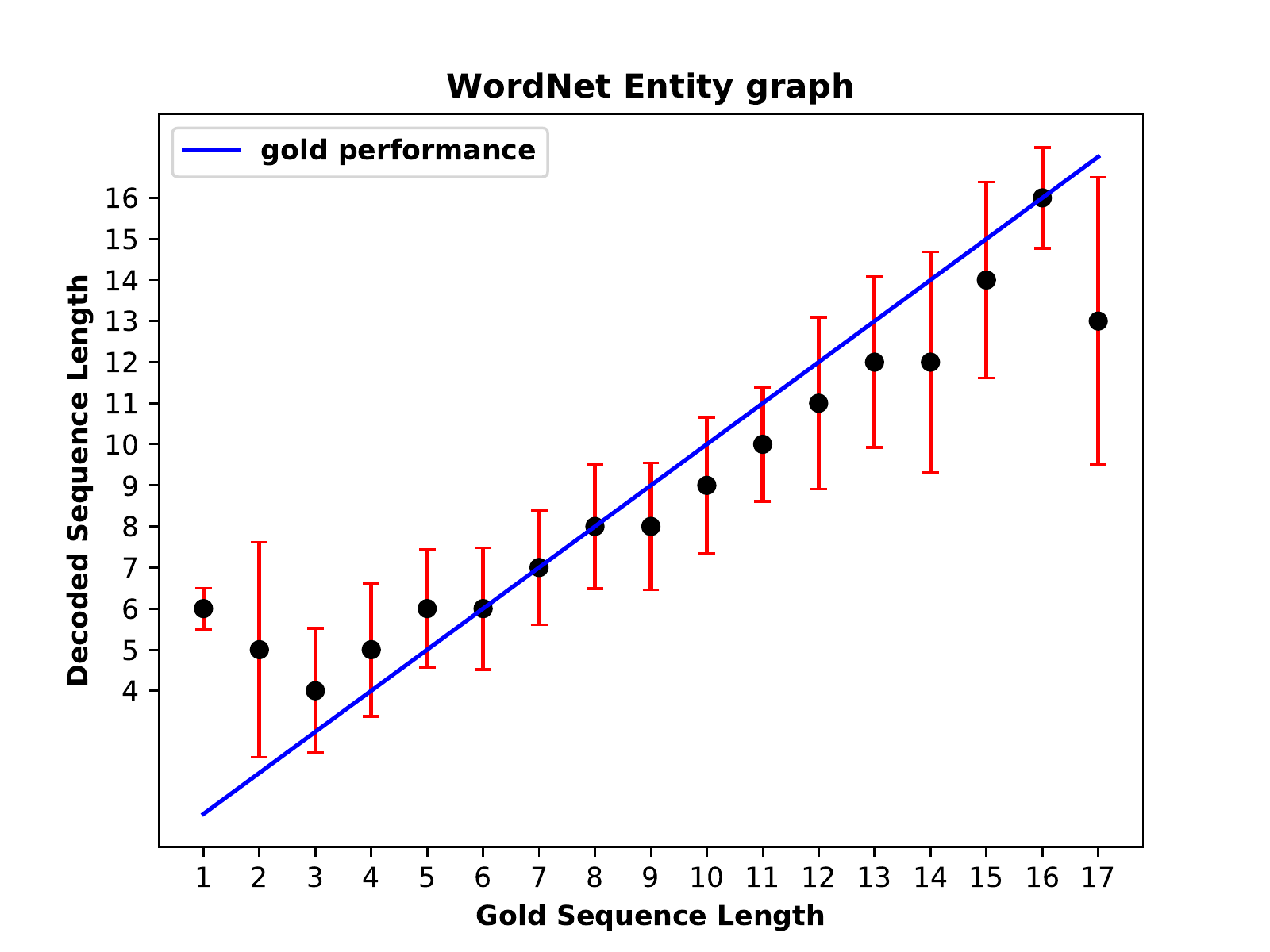}
\end{tabular}
\begin{tabular}{@{}c@{}}
    \includegraphics[height=2in]{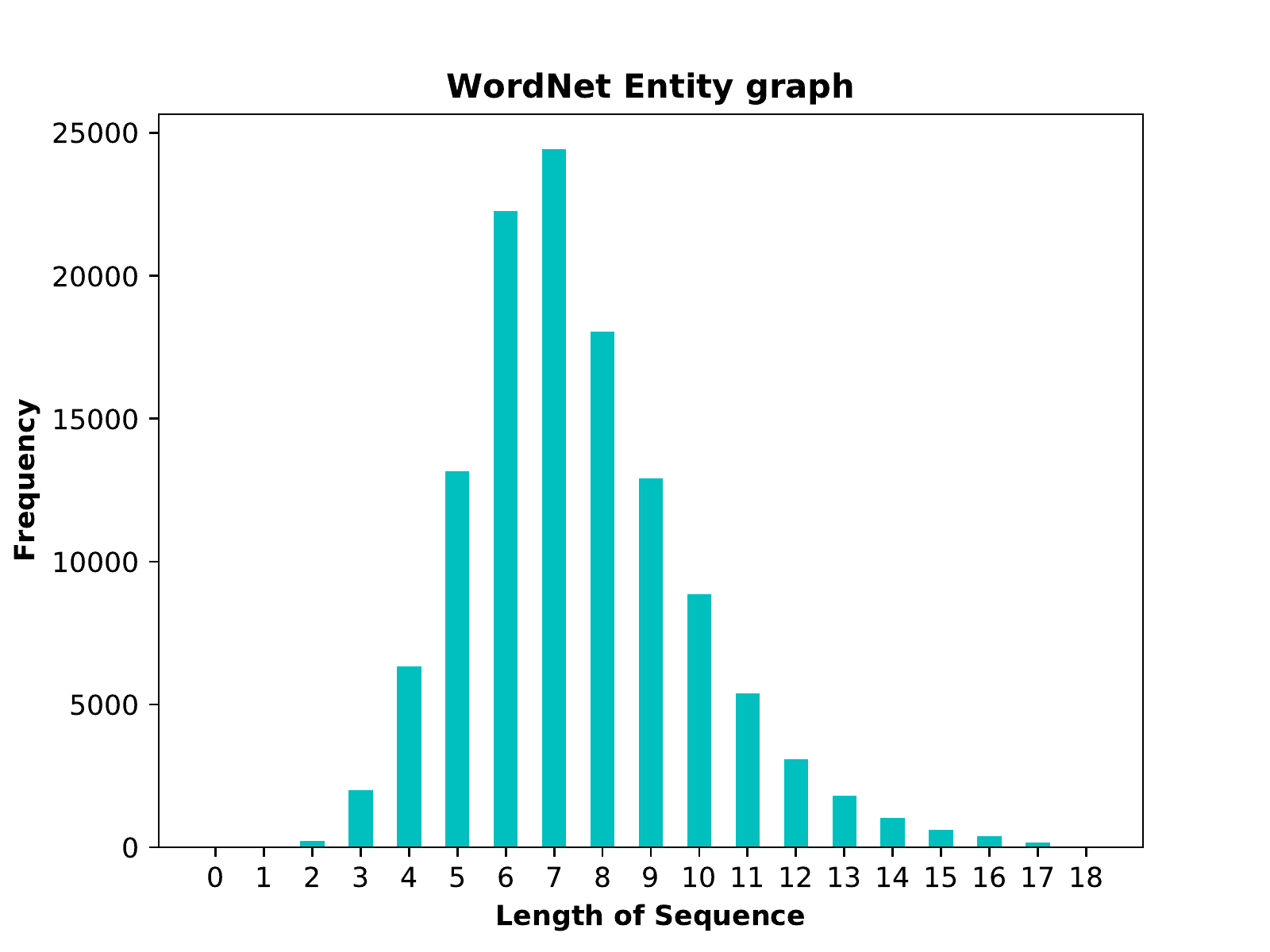}\\%
\end{tabular}    

    \caption{Continuation of Figure \ref{fig:ex3}. On the left graphs show: length of gold sequence vs mean length of decoded sequence on a test set; On the right graphs show: length of sequence vs length frequency on a training set.}%

    \label{fig:ex4}%
\end{figure*}

\end{document}